\documentclass[runningheads,a4paper]{llncs}

\usepackage{amssymb}
\usepackage{graphicx}
\usepackage{float}
\usepackage{array}
\usepackage{hyperref}
\usepackage{pbox}

\usepackage{url}

\begin{document}

Manchanda, P. (2013). Analysis of Optimization Techniques to Improve User Response Time of Web Applications and Their Implementation for MOODLE.\\
 \\
In Papasratorn, B.; Charoenkitkarn, N.; Vanijja, V.; Chongsuphajaisiddhi, V. (Eds.), Proceedings of 
the 6th International Conference, IAIT 2013, Bangkok, Thailand, December 12-13, 2013. 
\\
\\To appear in volume 0409 of Springer CCIS series. 
 \\The original final publication will be available at \url{www.springerlink.com} 

\title{Analysis of Optimization Techniques to Improve User Response Time of Web Applications and Their Implementation for MOODLE}
\titlerunning{Analysis of Optimization Techniques for Moodle LMS}
\author{Priyanka Manchanda}
\institute{Department of Computer Science and Information Technology,\\Jaypee Institute of Information Technology,\\Sector 128, Noida, UP - 201304, India\\
\url{pmanchanda1992@gmail.com}}
\maketitle

\begin{abstract}
Analysis of seven optimization techniques grouped under three categories (hardware, back-end, and front-end) is done to study the reduction in average user response time for Modular Object Oriented Dynamic Learning Environment (Moodle), a Learning Management System which is scripted in PHP5, runs on Apache web server and utilizes MySQL database software. Before the implementation of these techniques, performance analysis of Moodle is performed for varying number of concurrent users. The results obtained for each optimization technique are then reported in a tabular format. The maximum reduction in end user response time was achieved for hardware optimization which requires Moodle server and database to be installed on solid state disk.
\end{abstract}

\section{Introduction}
The Internet has seen a significant growth of web based applications over the last few years. These have now become an inseparable part of numerous industries like airline, banking, business, computer, education, financial services, healthcare, publishing and telecommunications. They are preferred because of their zero installation time (as they run on a browser), availability of centralised data, their global reach, and their availability (24 hours a day, 7 days a week). According to \cite{blog}, in June 2011 an average US user spent 74 minutes a day using web applications as compared to 64 minutes a day in June 2010.

In current scenario, improvement in the user response time is the most important issue for enhancing the performance of web applications. With reference to \cite{report}, a delay of one second in the performance of web applications can impact customer satisfaction by up to 16\%.

Web applications make use of a wide range of technologies including JavaScript, Apache, CSS, HTML, MySQL, PHP and protocols like HTTP headers. Optimizing the way they use these technologies can significantly improve user response time. Furthermore, the browser and hardware capabilities can be employed to reduce the user response time. 

Many research groups and authors have addressed this problem and reported their solutions. These include teams such as Yahoo Exceptional Performance Team \cite{yahoo}, book authors \cite{book_steve} and research papers \cite{paper}.

In this contribution, seven optimization techniques grouped under three categories are analysed. Further, implementation of these seven techniques is done for the Modular Object Oriented Distance Learning Environment (Moodle) \cite{moodle}. The efficiency of these techniques is studied by comparing the original and the improved average user response time.

\section{Performance Analysis of Moodle}
\subsection*{Performance Analysis for Varying Number of Concurrent Users}
\label{subsec:performance}
Moodle is a free source Learning Management System (LMS) which is used by thousands of educational institutions around the world to provide an organized interface for e-learning. As of June 2013, it has 83059 currently active sites that have been registered from  236 countries \cite{moodle_stats}. Moodle LMS is written in PHP and uses XHTML 1.0 Strict, CSS level 2 and JavaScript for its web user interface\cite{paper}.

With reference to \cite{moodle_install}, it has been reported that Moodle can support 50 concurrent users for every 1GB RAM.
An experiment was performed to verify this result.
\medskip \\
\textbf{Experimental Setup}
\\The experiment was perfomed on a machine with the following specifications:
\noindent
\medskip
\\{\it Hardware}: Intel\textregistered \thinspace  Core\texttrademark i5-2310 CPU @2.90GHz x 4 processor, 8GB Hard disk and 1GB RAM.
\\{\it Operating system}: Ubuntu 12.10
\\{\it Web server}: Apache v2.2.22 and PHP v5.4.6 for Moodle v2.5 for Ubuntu 12.10
\\{\it Database software}: MySQL v5.5.31 for Ubuntu 12.10
\medskip
\begin{itemize}
\item The experiment was performed using Apache JMeter 2.9, an open source load testing tool by the Apache Software Foundation \cite{jmeter}.
\item The test script was generated by using the JMeter Script Generator plugin for Moodle by James Brisland \cite{jmeter_plugin}.
\item The bandwidth of the network was set to 1024 kbps (1 Mbps) using JMeter.
\item The load testing of Moodle was done for a chat activity.
\item The sequence of pages visited on Moodle was :
\medskip
\\Login to site -$>$ View Course -$>$ View Chat page -$>$ View Chat window -$>$ Initialize Chat -$>$ Initialize Initial Update
\medskip
\item After initializing chat the following tasks were performed five times for each concurrent user : 
Post Chat Message -$>$ Initialize Update
\medskip
\item To test the performance of Moodle in the worst case scenario  the ramp-up period, that is the amount of time for creating the total number of threads, was set to zero so as to ensure immediate creation of all the threads by JMeter. 
\end{itemize}

\begin{table}[H]
\label{table:per}
\caption[Caption for LOF]{Average Response Time and Throughput for load testing Moodle \\ \footnotesize\centering on 1GB RAM and 8GB HARD DISK}
\centering
\begin{tabular}{ | c | c | c | }
\hline
\textit{\textbf{Number of Concurrent Users}} & \textit{\textbf{Average Response Time(s)}} & \textit{\textbf{Throughput (per min)}} \\ \hline
10 & 3.671 & 147.6 \\
20 & 8.874 & 129 \\
30 & 15.303 & 99.6 \\
40 & 129.786 & 16.8 \\
49 & 243.469 & 11.4 \\
50 & 364.480 & 7.8 \\
51 & Database Overload & Database Overload \\ \hline
\end{tabular}
\end{table}

\begin{figure}[H]
\centering
\includegraphics[height=4cm]{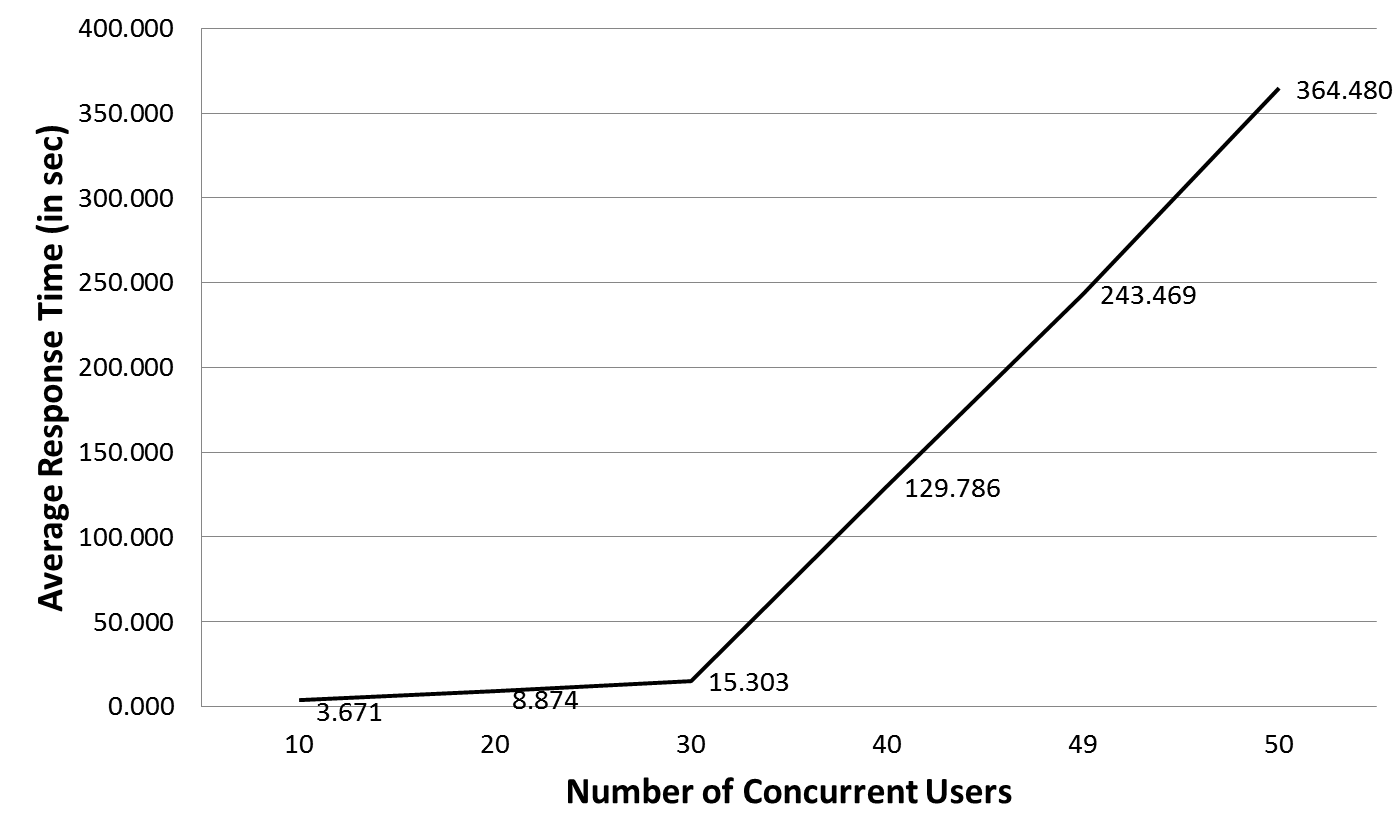}
\caption{Average Respose Time(in s) for Varying Number of Concurrent Users}
\label{fig:fig_perf1}
\end{figure}
\begin{figure}[H]
\centering
\includegraphics[height=4cm]{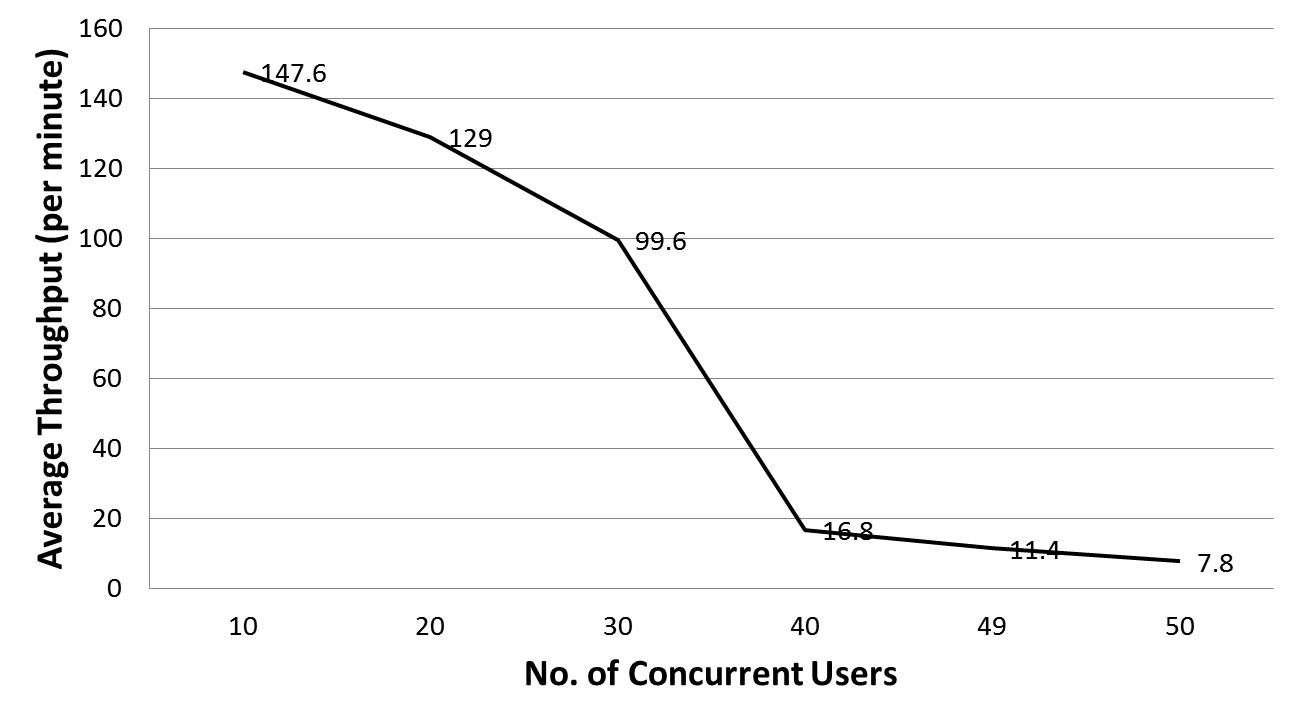}
\caption{Throughput(per minute) for Varying Number of Concurrent Users}
\label{fig:fig_perf2}
\end{figure}
While load testing Moodle for 51 concurrent users, it was observed that the connection to the database was aborted due to database overload and the testing process was killed by JMeter.

\section{Hardware Optimization}
\label{subsec:ssd}
\subsection*{Employing Solid State Disk}
The performance of the web applications can be highly enhanced by using a solid state disk drive to reduce the latency of the input and output operations carried out by the server.

A Solid State Disk, or SSD is a high performance plug and play data storage device which uses integrated circuit assemblies as memory to store data persistently \cite{wiki_ssd}. An SSD incorporates solid state flash memory and emulates a hard disk drive to store data \cite{paper_ssd}. However, unlike the traditional electromechanical disks like hard disk and floppy disks, an SSD is a flash-based and DRAM-based storage device which does not contain any moving parts \cite{article_ssd}.

An experiment was performed by replacing the Hard Disk Drive(HDD) of the Moodle Server with a 128GB Kingston Solid State Disk Drive.
\medskip \\
\textbf{Experimental Setup}
\\To conform to the experiment performed in section~\ref{subsec:performance} and to compare the performance of Moodle on HDD vs. SSD, the space allocated to Moodle server and database collectively was 8GB of 128GB SSD and the RAM size was limited to 1GB. The experiment was performed on a machine with following specifications:
\medskip
\\ \textbf{\textit{ Hardware:} Intel\textregistered \thinspace  Core\texttrademark i5-2310 CPU @2.90GHz x 4 processor,\\ 8GB Solid State disk and 1GB RAM.}
\\{\it Operating system}: Ubuntu 12.10
\\{\it Web server}: Apache v2.2.22 and PHP v5.4.6 for Moodle v2.5 for Ubuntu 12.10
\\{\it Database software}: MySQL v5.5.31 for Ubuntu 12.10
\\{\it Bandwidth}: 1024 Kbps (1 Mbps)
\medskip
\\The experiment was performed for the chat activity mentioned in  section~\ref{subsec:performance} using Apache JMeter 2.9.

\begin{figure}[H]
\centering
\includegraphics[height=3.6cm]{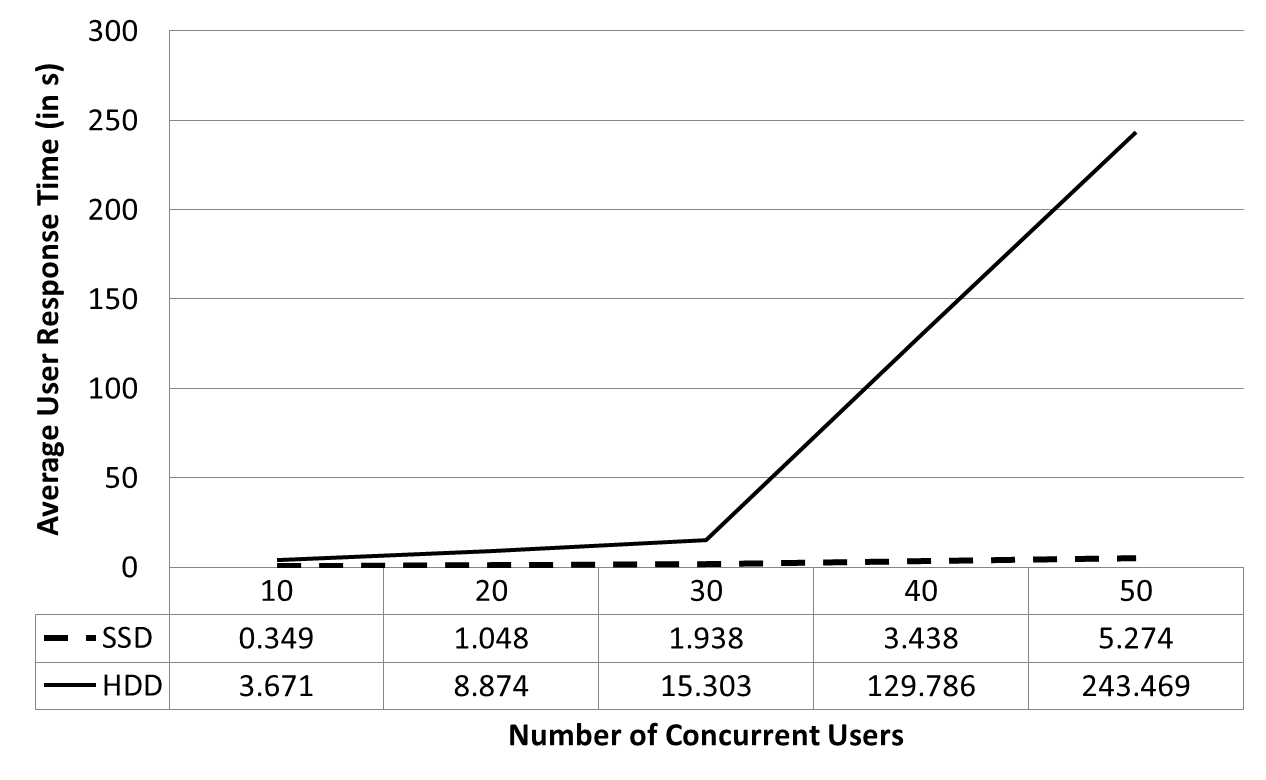}
\caption{Average user response time (in s) for Moodle on HDD vs. SSD}
\label{fig:fig_ssd}
\end{figure}

\begin{table}[H]
\begin{center}
 \caption{Average User Response Time on HDD vs SSD(in s)}
\begin{tabular}{|>{\centering\arraybackslash}p{3cm}|>{\centering\arraybackslash}p{3cm}|>{\centering\arraybackslash}p{3cm}|>{\centering\arraybackslash}p{3cm}|}
 \hline
 \textbf{No. of concurrent users} & \textbf{Average Response Time on HDD(s)} & \textbf{Average Response Time on SSD (s)} & \textbf{Reduction in Response time \%} \\
 \hline
10 & 3.671 & 0.349	& 90.49\\
\hline
20 & 8.874 & 1.048 & 88.19\\
\hline
30 & 15.303 & 1.938 & 87.34\\
\hline
40 & 129.786 & 3.438 & 97.35\\
\hline
50 & 364.480 & 5.274 & 97.83\\
\hline
60 & Database Overload & 5.97& -\\
\hline
70 & Database Overload & 6.492	& -\\
\hline
80 & Database Overload & 8.009	& -\\
\hline
90 & Database Overload & 8.085 & -\\	
\hline
100 & Database Overload	& 9.797 &-\\
\hline
110 & Database Overload & 13.759 &-\\
\hline
120 & Database Overload & 16.828 & -\\
\hline
130 & Database Overload & 22.991 & -\\
\hline
140 & Database Overload	& 30.187& -\\
\hline
150 & Database Overload	& 36.119 & -\\
\hline
151 & Database Overload	& 39.141 & -\\
\hline
152 & Database Overload & Database Overload & -\\
 \hline
\end{tabular}
\end{center}
\end{table}

From Table 2, it is concluded that the number of concurrent users supported by Moodle installed on SSD for 1 GB RAM is increased to \textbf{151} as compared to \textbf{50} concurrent users for Moodle installed on HDD with 1 GB RAM. Also, there is a reduction of \textbf{87\% to 98\%} in average user response time after installing Moodle server and database on SSD.

\section{Back-End Optimization}
\subsection*{Switching to LNMP Stack from LAMP Stack}
\label{sec:back}
The Moodle web application runs on the LAMP stack which is a software bundle comprising of Linux based operating system, Apache HTTP server, MySQL database software and PHP object oriented scripting language. LNMP stack is almost similar to LAMP, except the change of web server from Apache to Nginx. 

Apache is a process-based server, while nginx is an event-based asynchronous web server and is more scalable than Apache. In Apache, each simultaneous connection requires a thread which incurs significant overhead whereas nginx  is event-driven and handles requests in a single (or at least, very few) threads \cite{wiki_apache_nginx}.

The performance of Moodle or any web application that runs on Apache and frequently encounters heavy load, can be boosted by replacing Apache by Nginx. An experiment was performed to compare the performance of Moodle installed on Apache vs. Nginx web server. 
\medskip \\
\textbf{Experimental Setup}
\\Since it was observed in Section~\ref{subsec:ssd} that the performance of Moodle is highly enhanced by installing it on SSD, the experiment was performed on a machine with Moodle installed on 128 GB Solid State Disk and 4GB RAM.

All the other specifications (Operating system, Database software, Web server and Bandwidth)  of the machine were kept same as  in section~\ref{subsec:ssd}.
The experiment was performed using Apache BenchMark 2.4 \cite{ab} for Moodle's login page.

From Table 3, it is observed that there is a reduction of \textbf{24\% to 34\%} in average user response time after installing Moodle on Nginx v1.4.1 web server.

\begin{table}[H]
\begin{center}
 \caption{Average User Response Time on Apache vs Nginx(in s) Web Server}
\begin{tabular}{|>{\centering\arraybackslash}p{3cm}|>{\centering\arraybackslash}p{3cm}|>{\centering\arraybackslash}p{3cm}|>{\centering\arraybackslash}p{3cm}|}
 \hline
 \textbf{No. of concurrent users} & \textbf{Average Response Time on Apache(s)} & \textbf{Average Response Time on Nginx (s)} & \textbf{Reduction in Response time \%} \\
 \hline
50 & 2.209 & 1.652	& 25.22\\
\hline
100 & 4.505 & 3.359 & 25.43\\
\hline
150 & 6.098 & 4.630 & 24.07\\
\hline
200 & 8.192 & 5.408 & 33.98\\
\hline
250 & 10.729 & 7.156 & 33.30\\
\hline
\end{tabular}
\end{center}
\end{table}

\begin{figure}[H]
\centering
\includegraphics[height=3cm]{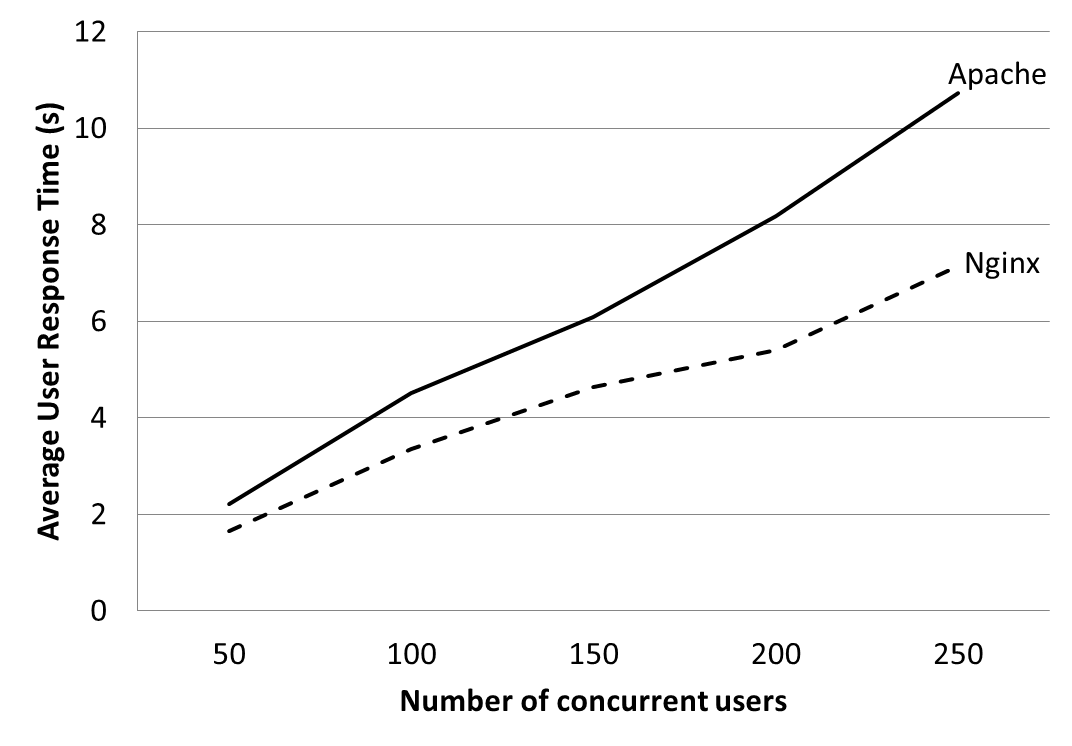}
\caption{Average user response time (in s) for Moodle on Apache vs. Nginx Web Server}
\label{fig:fig_nginx}
\end{figure}

\section{Front-End Optimization}
For any web application, only 10\% to 20\% of the end user response time is spent downloading the HTML document from the web server to the client's browser. The other 80\% to 90\% is spent in performing the front end operations, i.e., in downloading the other components of web page \cite{book_steve}.

A set of specific rules for speeding up the front – end operations carried by a web application is presented in Ref.~\cite{book_steve}.  Five of the most efficient techniques which showed significant reduction in user response time for Moodle Learning Management System are described in this section.

\subsection{Browser Caching by Using Far Future Expires Header}
\label{subsec:caching}
Browsers and proxies use cache to reduce the number and size of the HTTP requests thereby speeding up the web applications. A first-time visitor may have to make several HTTP requests, but by using a Far Future Expires header the developer can significantly improve the performance of web applications for returning visitors. A server uses the Expires header in HTTP response to inform the client that it can use the current copy of a component until the specified time \cite{book_steve}.

Moodle sends requests with an Expires Header which is set in past \textbf{(20th Aug 1969 09:23 GMT)}. An experiment was performed by changing it to future date of \textbf{16th Apr 2015 20:00 GMT}. Also max-age directive was used in Cache control header so as to set the cache expiration window to 10 years in future and the pragma header was unset to enable caching.
\\ \\Given below are the lines which were added to the headers.conf file of Apache2 Web Server:

\noindent
\begin{verbatim}
<FilesMatch ".(ico|pdf|flv|jpg|jpeg|png|gif|js|css|swf|php|html)$">
Header set Expires "Thu, 16 Apr 2015 20:00:00 GMT"
Header set Cache-Control " max-age=315360000"
Header unset Pragma
</FilesMatch>
\end{verbatim}
The experimental setup is the same as section~\ref{subsec:ssd} and the experiment was performed using Apache Jmeter 2.9. From Table 4, it is observed that there is a reduction of \textbf{70\% to 80\%} in average user response time after implementing Far Future Expires Header Optimization Technique.

\begin{table}[H]
\begin{center}
 \caption{Average user response time (in s) with and without caching for 10 iterations}
\begin{tabular}{|>{\centering\arraybackslash}p{3cm}|>{\centering\arraybackslash}p{3cm}|>{\centering\arraybackslash}p{3cm}|>{\centering\arraybackslash}p{3cm}|}
 \hline
 \textbf{No. of concurrent users} & \textbf{Average Response Time Without Expires Header(s) (no caching)} & \textbf{Average Response Time with Expires Header(s) (caching)} & \textbf{Reduction in Response time \%} \\
 \hline
10 & 0.625 & 0.144	& 76.96\\
\hline
20 & 1.839 & 0.408 & 77.81\\
\hline
40 & 5.061 & 1.210 & 76.09\\
\hline
60 & 7.086 & 1.778 & 74.91\\
\hline
80 & 8.124 & 2.426 & 70.14\\
\hline
100 & 9.882 & 3.071 & 68.92\\
\hline
\end{tabular}
\end{center}
\end{table}

\subsection{Reduce DNS Lookups}
\label{subsec:dns}

The Internet uses IP addresses to find webservers. Before establishing a network connection to a web server, the browser must resolve the hostname of the web server to an IP address by using Domain Name Systems (DNS). The latency introduced due to DNS lookups can be minimized if the DNS resolutions are cached by client's browser \cite{book_steve}. The response time for Moodle's login page of Institutional Moodle websites of 13 universities situated in six continents of the world was recorded for two cases: With DNS Cache and Without DNS Cache.\\The experiment was performed for a client located in IIT Bombay, India with 128GB SSD, 4GB RAM, Intel\textregistered \thinspace  Core\texttrademark i5-2310 CPU @2.90GHz x 4 processor and 2 Mbps average download speed.

\begin{table}[H]
\begin{center}
 \caption{Average user response time (in s) With and Without DNS Cache for 1 user}
\begin{tabular}{|>{\centering\arraybackslash}p{1.5cm}|>{\centering\arraybackslash}p{1.5cm}|>{\centering\arraybackslash}p{4.3cm}|>{\centering\arraybackslash}p{1.5cm}|>{\centering\arraybackslash}p{1.5cm}|>{\centering\arraybackslash}p{1.6cm}|}
 \hline
 \textbf{Continent} & \textbf{Country} & \textbf{University} & \textbf{Response time With DNS Cache(s)} & \textbf{Response time Without DNS Cache(s)} & \textbf{Reduction in Response time(\%)} \\
 \hline
Asia & India & IIT, Bombay \cite{iitb} & 2.357 & 1.426 & 39.50\\
\hline
Asia & India & IIT, Madras \cite{iitm} & 2.516 & 1.612 & 35.93\\
\hline
Asia & Singapore & SIM University \cite{sim} & 1.381 & 1.055 & 23.61\\
\hline
Asia & Japan & Sojo University, Kumamoto \cite{sojo} & 6.223 & 3.116 & 49.93\\
\hline
Europe & Spain & Graduate School of Management, Barcelona \cite{spain} & 3.138 & 1.813 & 42.22\\
\hline
Europe & UK & University of Nottingham \cite{uk} & 4.174 & 2.041 & 51.10\\
\hline
North America & US & UCLA, California \cite{ucla} & 4.600 & 3.657 & 20.50\\
\hline
South America & Argentina & Pontifical Catholic University of Argentina, Buenos Aires \cite{arg}& 2.534 & 1.710 & 32.52\\
\hline
South America & Colombia & University of Grand Colombia, Bogotá, D.C. \cite{colombia} & 2.341 & 1.438 & 38.57\\
\hline
Africa & Egypt & Oriflame University \cite{egypt} & 5.497 & 4.288 & 21.99\\
\hline
Africa & South Africa & Virtual Academy of South Africa \cite{sa}& 4.936 & 2.588 & 47.57\\
\hline
Australia & Australia & Australian National University \cite{aun}& 4.525 & 3.559 & 21.35\\
\hline
Australia & Australia & Monash University \cite{monash} & 4.947 & 4.141 & 16.29\\
\hline
\end{tabular}
\end{center}
\end{table}

From Table 5 it is concluded that there is a reduction of \textbf{16\% to 51\% depending on the geographical location} of Moodle server, if the resolved hostname for a web page is found in DNS cache.

Another experiment was carried out on the same client to compare the performance of Moodle by changing the number of DNS cache entries, DNS cache expiration period and HTTP keep alive timeout for Mozilla Firefox 21.0 browser. The following three scenarios were tested for 100 iterations of Moodle's login page of  Moodle websites of six universities situated in six continents of the world using iMacros 9.0 Firefox extension \cite{imacros} and HttpFox addon for Firefox \cite{httpfox}.
\noindent

\begin{verbatim}
Scenario 1 (S1):  
DNS Cache Entries = 20
DNS Cache Expiration Period = 60 seconds
HTTP Keep Alive Timeout = 115 seconds
 
Scenario 2 (S2):
DNS Cache Entries = 512
DNS Cache Expiration Period = 3600 seconds
HTTP Keep Alive Timeout = 115 seconds
 
Scenario 3 (S3):
DNS Cache Entries = 512
DNS Cache Expiration Period = 3600 seconds
HTTP Keep Alive Timeout = 0 second
 \end{verbatim}

\begin{table}[H]
\begin{center}
 \caption{Average user response time (s) for 1 user, 100 iterations for above Scenarios}
\begin{tabular}{|>{\centering\arraybackslash}p{1.7cm}|>{\centering\arraybackslash}p{2.5cm}|>{\centering\arraybackslash}p{1.5cm}|>{\centering\arraybackslash}p{1.5cm}|>{\centering\arraybackslash}p{2cm}|>{\centering\arraybackslash}p{1.5cm}|>{\centering\arraybackslash}p{2cm}|}
 \hline
 \textbf{Continent} & \textbf{University} &  \textbf{Response time for S1} & \textbf{Response time for S2} & \textbf{Difference (s) between S1 \& S2} & \textbf{Response time for S3} &  \textbf{Difference(s) between S2 \& S3} \\
 \hline
North America & UCLA, USA \cite{ucla} & 173.984 & \textbf{169.284} & 4.7 & 178.69 & 9.406\\
\hline
Asia &  IIT, Madras, India \cite{iitm} & 108.93 & \textbf{105.677} & 3.253 & 110.548 & 4.871\\
\hline
Australia &  Australian National University \cite{aun} & 347.361 & \textbf{344.961} & 2.400 & 354.336 & 9.375\\
\hline
Africa & Oriflame University, Egypt \cite{egypt} & 244.035 & \textbf{240.246} & 3.789 & 256.08 & 15.834\\
\hline
Europe & University of Nottingham, UK \cite{uk} & 153.71 & \textbf{150.213} & 3.497 & 156.76 & 6.547\\
\hline
South America & University of Grand Colombia, Colombia \cite{arg} & 142.241 & \textbf{135.908} & 6.333 & 146.763 & 10.855\\
\hline
\end{tabular}
\end{center}
\end{table}

From Table 6, it is observed that the end user response time is minimum under Scenario 2. Hence, it can be concluded that the performance of a web application can be enhanced by reducing DNS Lookups, which was achieved by:
\begin{itemize}
 \item Increasing the number of DNS cache entries, 
 \item Increasing DNS expiration period, and
 \item Using a Network that supports HTTP keep-alive mechanism
\end{itemize}

\subsection{Gzip Components}
\label{subsec:gzip}
Gzip compression of web pages can significantly minimize the latency introduced due to transfer of the web page files from web server to client's browser. Starting with HTTP/1.1, web clients indicate support for compression with the Accept-Encoding header in the HTTP request \cite{book_steve}.
\begin{verbatim}
Accept-Encoding: gzip, deflate
\end{verbatim}
After the web server sees this header, it compresses the response using one of the methods listed by the client. The web server uses the Content-Encoding header in the response to inform the client about the compressed response \cite{book_steve}.
\begin{verbatim}
Content-Encoding: gzip
\end{verbatim}
An experiment was performed on the client mentioned in section~\ref{subsec:dns} for Moodle installed on the machine with specifications as mentioned in section~\ref{sec:back} using Web Developer Extension for Mozilla Firefox 21.0 \cite{web_dev}.
\begin{table}[H]
\begin{center}
 \caption{Response Size of Moodle pages with and without compression of components}
\begin{tabular}{|>{\centering\arraybackslash}p{2.8cm}|>{\centering\arraybackslash}p{2cm}|>{\centering\arraybackslash}p{3cm}|>{\centering\arraybackslash}p{2.5cm}|>{\centering\arraybackslash}p{2cm}|}
 \hline
 \textbf{Moodle Page} & \textbf{No. of Files Requested} &  \textbf{Response Size without Compression(KB)} & \textbf{Response size with Compression(KB)} & \textbf{Reduction in Response Size (\%)}  \\
 \hline
Index & 42 & 926 & 215 & 76.78\\
\hline
Login & 13 & 597 & 138 & 76.88\\
\hline
View Course & 42 & 804 & 187 & 76.74\\
\hline
\vbox{View Forum}\pbox{2.8cm}{(with 1 post)} & 41 & 802 & 187 & 76.68\\
\hline
\vbox{View Blog}\pbox{2.8cm}{(with 10 posts)} & 35 & 889 & 218 & 75.48\\
\hline
View Calendar & 42 & 861 & 207 & 75.96\\
\hline
\vbox{View Participants}\pbox{2.8cm}{(20 per page)} & 40 & 806 & 188 & 76.67\\
\hline 
1 page quiz with 5 questions & 49 & 847 & 198 & 76.62\\
\hline
View Assignments & 43 & 804 & 187 & 76.74\\
\hline
\end{tabular}
\end{center}
\end{table}
It is observed that Gzip compression reduces the response size by \textbf{75\%\mbox{-}77\%}. 

\subsection{Deactivate ETags}
\label{subsec:etag}
Entity tags (ETags) are used by web servers and browsers to determine whether the component in the browser's cache 

An experiment was performed on the client mentioned in section~\ref{subsec:dns} for Moodle installed on the machine with specifications as mentioned in section~\ref{sec:back} using Firebug Extension 1.11.4 for Mozilla Firefox 21.0 \cite{firebug}. Moodle uses Etags for its style sheets, scripts and images. It was observed that deactivating Etags reduces the response 

There is a reduction of  \textbf{an average of 1737 bytes}  in response header size after deactivating the ETags.

\begin{table}[H]
\begin{center}
 \caption{Response Header Size of Moodle pages with activated and deactivated ETags}
\begin{tabular}{|>{\centering\arraybackslash}p{2.3cm}|>{\centering\arraybackslash}p{3cm}|>{\centering\arraybackslash}p{2.5cm}|>{\centering\arraybackslash}p{2.5cm}|>{\centering\arraybackslash}p{1.8cm}|}
 \hline
 \textbf{Moodle Page} & \textbf{Total No. of CSS style sheets, JS Files and Images Requested} &  \textbf{Response Header size with activated Etags(Bytes)} & \textbf{Response Header Size with deactivated Etags(Bytes)} & \textbf{Reduction in Response Size (Bytes)}  \\
 \hline
Index & 40 & 20357 & 18454 & 1921\\
\hline
Login & 11 & 5695 & 5172 & 523\\
\hline
View Course & 42 & 21364 & 19345 & 2019\\
\hline
1 page quiz with 5 questions & 49 & 19850 & 17978 & 1872\\
\hline
View Assignments & 39 & 24978 & 22628 & 2350\\
\hline
\textbf{Average} & & & & \textbf{1737} \\
\hline
\end{tabular}
\end{center}
\end{table}

\subsection{Optimize AJAX}
AJAX (Asynchronous JavaScript and XML) is a collection of technologies, primarily JavaScript, CSS, DOM and asynchronous data retreival which is used to exchange data with a server, and update parts of a web page \mbox{-} without reloading the whole page. Though AJAX allows the server to provide instantaneous feedback to the user, it does not guarantee that the user won't have to wait for the asynchronous JavaScript and XML responses. The performance of the web application can be improved by optimizing the AJAX requests. The techniques mentioned in section~\ref{subsec:caching}, \ref{subsec:dns} and \ref{subsec:gzip} are collectively used to optimize the ajax components of Moodle.

The AJAX components are made cacheable by modifying the expires header which is defined in \textbf{OutputRenderers.php} file located in \textbf{lib} directory of main Moodle directory. An experiment was performed on the client mentioned in section~\ref{subsec:dns} for Moodle installed on the machine with specifications as mentioned in section~\ref{sec:back} using Firebug Extension 1.11.4 for Mozilla Firefox 21.0 \cite{firebug}.
\medskip
\\ \textit{Modifications made to expires header in OutputRenderers.php file}
\begin{verbatim}
Default:
@header('Expires: Sun, 28 Dec 1997 09:32:45 GMT'); Line 3345

Modified:
@header('Expires: Sun, 28 Dec 2020 09:32:45 GMT'); Line 3345
\end{verbatim}

\begin{table}[H]
\begin{center}
 \caption{Average User Response Time (s) before and after optimizing AJAX }
\begin{tabular}{|>{\centering\arraybackslash}p{2.3cm}|>{\centering\arraybackslash}p{3cm}|>{\centering\arraybackslash}p{2.5cm}|>{\centering\arraybackslash}p{2.5cm}|>{\centering\arraybackslash}p{1.8cm}|}
 \hline
 \textbf{Activity} & \textbf{Response Time before optimizing AJAX(s)} &  \textbf{Response Time after optimizing AJAX(s)} & \textbf{Reduction in Response Time(\%)}  \\
 \hline
Drag and Drop Sections & 309 & 227 & 26.54 \\
\hline
Drag and Drop Activities & 2.17 & 1.62 & 25.35 \\
\hline
Drag and Drop Files & 201 & 175 & 12.94\\
\hline
Drag and Drop Blocks & 440 & 354 & 19.55\\
\hline
AJAX Chat Box & 36 & 24 & 33.33\\
\hline
\textbf{Average} & & & \textbf{23.54}\\
\hline
\end{tabular}
\end{center}
\end{table}

There is a reduction of an average of 23.54\% in user response time after optimizing the AJAX components.

\section{Conclusion}
In this presented paper, seven methods to optimize web applications have been analysed and tested for Moodle LMS. These methods can be further used to optimize other essential web applications including webmail, online retail sales, online auctions, wikis and e-learning.
It is observed that Moodle shows faster response time under heavy traffic, if it is loaded on a solid state disk. This technique can be used to scale high traffic web applications.

The caching mechanism can be used by the client’s browser to optimize the front-end operations that can reduce the end-user response time by up to 80\%. This mechanism can be used for content that changes infrequently, that is, application's static assets like graphics, style sheets and scripts. In addition to application's static assets, DNS resolutions can be cached by client's browser and can reduce the end-user response time by up to 50\%.

High traffic web applications can employ more than one server to share the load. In such a scenario, Etags become invalid and deactivating them can further enhance the performance of the web application.

The seven web optimization techniques discussed in this paper were successfully tested for the Moodle LMS which showed a maximum reduction of 98\% in average user response time by using the hardware optimization technique used in  (Section~\ref{subsec:ssd}). These best practices can be further applied to a novel or existing web application to improve its performance by reducing end user response time and thereby increasing the number of concurrent users and throughput.

\section*{Acknowledgement}
The authors would like to thank the members of Department of Computer Science, Indian Institute of Technology, Bombay, India for their kind support and encouragement.

\end{document}